\documentclass[pdflatex,sn-mathphys]{sn-jnl}% Math and Physical Sciences Reference Style
\jyear{2021}%

\theoremstyle{thmstyleone}%
\theoremstyle{thmstyletwo}%

\theoremstyle{thmstylethree}%

\newcommand{\ie}{i.\,e., }
\newcommand{\eg}{e.\,g., }

\usepackage{paralist}
\usepackage{multirow}
\usepackage{caption, subcaption, booktabs}
\usepackage{mathtools}
\usepackage{amssymb}

\usepackage[superscript,biblabel,nomove ]{cite}

\begin{document}

\title[Propagating Variational Model Uncertainty for Bioacoustic Call Label Smoothing]{Propagating Variational Model Uncertainty for Bioacoustic Call Label Smoothing}


\author*[1]{\fnm{Georgios} \sur{Rizos}}\email{georgios.rizos12@imperial.ac.uk}

\author[2]{\fnm{Jenna} \sur{Lawson}}\email{j.lawson17@imperial.ac.uk}

\author[3]{\fnm{Simon} \sur{Mitchell}}\email{s.mitchell@kent.ac.uk}

\author[1]{\fnm{Pranay} \sur{Shah}}\email{pranay.shah18@imperial.ac.uk}

\author[1]{\fnm{Xin} \sur{Wen}}\email{xin.wen20@imperial.ac.uk}

\author[2]{\fnm{Cristina} \sur{Banks-Leite}}\email{c.banks@imperial.ac.uk}

\author[2]{\fnm{Robert} \sur{Ewers}}\email{r.ewers@imperial.ac.uk}

\author[1,4]{\fnm{Bj{\"o}rn W.} \sur{Schuller}}\email{bjoern.schuller@imperial.ac.uk}

\affil*[1]{\orgdiv{GLAM -- Group on Language, Audio, \& Music, Department of Computing}, \orgname{Imperial College London}, \orgaddress{\country{UK}}}

\affil[2]{\orgdiv{Department of Life Sciences}, \orgname{Imperial College London}, \orgaddress{\country{UK}}}

\affil[3]{\orgdiv{DICE -- Durrell Institute of Conservation and Ecology}, \orgname{University of Kent}, \orgaddress{\country{UK}}}

\affil[4]{\orgdiv{EIHW -- Chair of Embedded Intelligence for Health Care and Wellbeing}, \orgname{University of Augsburg}, \orgaddress{\country{Germany}}}

\maketitle

\section*{Highlights}
\label{sec:highlights}

\begin{itemize}
    \item Sample-free, Bayesian attentive ResNet with squeeze-and-excitation
    \item Uncertainty based, data-specific label smoothing
    \item Bioacoustic call detection on two datasets, one of which is introduced here
    \item Propagated uncertainty should be used to weigh label smoothing
\end{itemize}

\section*{The Bigger Picture}
\label{sec:thebiggerpicture}

Neural networks that accompany their predictions with uncertainty values can foster trust in artificial intelligence based decision-making. We leverage the potential of Bayesian deep learning by using its efficiently propagated uncertainty as a guide for adaptive label smoothing regularisation. This methodology is potentially transferrable to other audio classification problems, like voice-based pulmonary disease detection and affective computing, but even further, to textual, video, and graph data. We focus on wildlife call detection, a task that can upscale conservation experts' labour in annotating automated field recordings made by on-site passive acoustic monitoring devices. Such automated analysis can both reduce human presence in the field, and enhance policy-making pertaining to conservation.

% \ie without computationally costly Monte Carlo sampling of weights,

\section*{Summary}
\label{sec:summary}

% What is the problem.
% What have others done and why have failed at it
% What is our insight

% What do we propose

% A sentence of results.

% Broader significance of the work.

We focus on using the predictive uncertainty signal calculated by Bayesian neural networks to guide learning in the self-same task the model is being trained on. Not opting for costly Monte Carlo sampling of weights, we propagate the approximate hidden variance in an end-to-end manner, throughout a variational Bayesian adaptation of a ResNet with attention and squeeze-and-excitation blocks, in order to identify data samples that should contribute less into the loss value calculation. We, thus, propose uncertainty-aware, data-specific label smoothing, where the smoothing probability is dependent on this epistemic uncertainty. We show that, through the explicit usage of the epistemic uncertainty in the loss calculation, the variational model is led to improved predictive and calibration performance. This core machine learning methodology is exemplified at wildlife call detection, from audio recordings made via passive acoustic monitoring equipment in the animals' natural habitats, with the future goal of automating large scale annotation in a trustworthy manner.

\section*{Graphical Abstract}
\label{sec:graphicalabstract}

\begin{figure}[h]%
\centering
\includegraphics[width=0.45\textwidth]{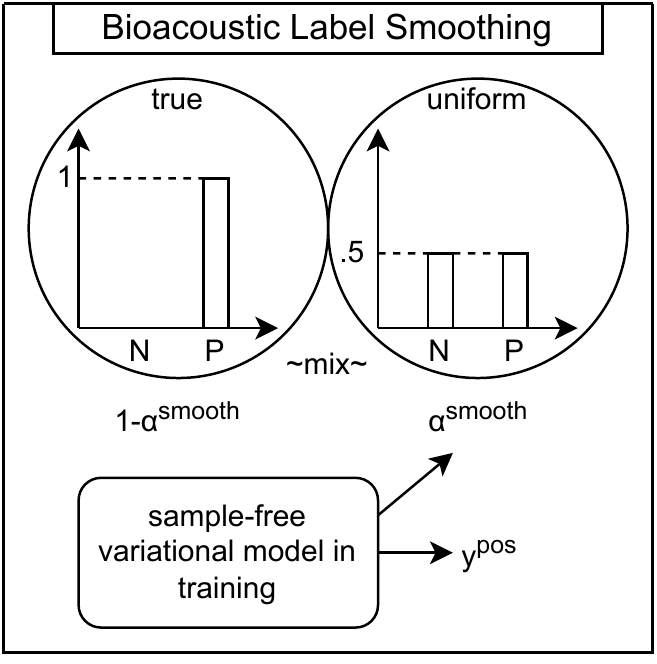}
% \caption{This is a widefig. This is an example of long caption this is an example of long caption  this is an example of long caption this is an example of long caption}
\label{fig:graphicalabstract}
\end{figure}

\section*{Keywords}
\label{sec:keywords}

\begin{inparaenum}[]
  \item variational-Bayesian-deep-learning,
  \item uncertainty-propagation,
  \item adaptive-label-smoothing,
  % \item distributional-max-pooling,
  \item epistemic-uncertainty,
  \item calibrated-deep-learning,
  \item bioacoustics,
  \item wildlife-call-detection,
  \item passive-acoustic-monitoring,
  \item machine-listening.
\end{inparaenum}

\section*{Data Science Maturity}
\label{sec:datasciencematurity}

DSML 3: Development/pre-production: Data science output has been rolled out/validated across multiple domains/problems

\section{Introduction}
\label{sec:introduction}

Effective wildlife monitoring that can guide action to ameliorate the effects of the global biodiversity crisis poses an enormous scalability challenge \cite{witmer2005wildlife,tuia2022perspectives}. To that end, the combination of audio sensing infrastructure \cite{turner2014sensing} and Deep Learning (DL) methods for bioacoustic data modelling \cite{stowell2022computational} holds great potential as a solution. The use of Passive Acoustic Monitoring (PAM) hardware \cite{turner2014sensing} allows for an automated, continuous monitoring solution that minimises the duration of human presence in the field, and, thus, the impact it can have on the behaviour of the animals. The recording, also, no longer needs to be limited to how much experts can reasonably listen, leading to great scalability both spatially and temporally. Further, DL for bioacoustics offers the possibility of distilling the detection and categorisation experience of ecology experts into an artificial intelligence (AI) computational model that can automate and expedite relevant labour, alleviating spurious annotation errors (as DL is known to be capable of doing \cite{veit2017learning}), such that the time of experts can be invested in a more fruitful manner. This scaled-up data enrichment can improve contributions to conservation- and ecology-related policy-making \cite{arroyo2014landscape}.

The modelling capacity of DL architectures that saw success in the visual classification domain, like Residual Networks (ResNets) \cite{he2016deep}, Squeeze-and-Excitation (SE) blocks \cite{hu2018squeeze}, and attention mechanisms for sequence pooling \cite{bahdanau2015neural,luong2015effective}, have been shown to be transferrable to the acoustic domain as well \cite{hershey2017cnn,naranjo2020acoustic,zhang2019attention}, with great advances in Acoustic Event Detection (AED) \cite{kong2020panns}. Adaptations and improvements also exist in the bioacoustics domain, in tasks like call detection \cite{rizos2021multi}, cross- \cite{shiu2020deep}, or within-species \cite{hantke2018my} call type classification, and individual identification \cite{oikarinen2019deep}; such liberally selected applications exist across a wide range in taxonomy, \eg on primates \cite{clink2019gibbonfindr,tzirakis2020computer,rizos2021multi}, whales \cite{shiu2020deep}, birds \cite{goeau2016lifeclef,rovithis2021towards}.

It is important, then, that the predictions made by the AI model are understood and trusted. Unfortunately, during this near-decade of DL advancement, a fixation by the AI community towards deeper and more complicated architectures, as well as on traditional prediction performance evaluation measures, has led to an insidious DL model behaviour manifesting: overconfident predictions \cite{guo2017calibration}, \ie predictions made at a probability nearing $1$, regardless of whether they are correct or not. Floods of overconfidently predicted misclassifications, with downstream software modules or policy-makers making catastrophic decisions due to this lack of introspective filtering, can foster deep mistrust in AI \cite{guo2017calibration,tomsett2020rapid,tomani2021towards}, something that has also been noted with respect to bioacoustics \cite{stowell2022computational}. However, early prediction calibration fixes \cite{guo2017calibration} are based on learning a transformation of the model outputs that require the existence of a validation set of labels, something that cannot be safely assumed in general.
%PS: 'cannot be safely assumed *in* general'? - Unless I misunderstood the meaning.

% We believe that the model should be able to understand when it is uncertain about a particular sample, and to express that with \textit{well-calibrated} predictions.

A means of designing DL models with an intrinsic capacity for knowing when they do not
%BS: please avoid colloquial style such as "don't" :)
know something -- expressed quantitatively as a measure of predictive uncertainty -- is (approximate) Bayesian inference; specifically, Bayesian Neural Networks (BNNs) \cite{mackay1992bayesian}, which have been shown to naturally offer better calibrated outputs \cite{maddox2019simple}. BNNs employ distributional weight parameters, of which the posterior distributions are calculated via Bayes' law, and dependent on the observed training set and a prior distribution assumption \cite{mackay1992bayesian}. Since, however, the integration for these posteriors is intractable, marginalising the weights in order to get the statistical distribution of the outputs is commonly based on Monte Carlo (MC) sampling. This means that if one wishes to propagate model parameter (epistemic) uncertainty in an end-to-end manner, \ie from the input (if stochastic) to the model output, one has to use $K$ MC samples, something that increases the computational load by $K$.

%PS: Above, I think it might be 'marginalising out the weights' as opposed to 'marginalising the weights'? 

% Mention variational. Is the thing about marginalisation correct? Do we do this because of variational?

One approach for \textit{sample-free uncertainty propagation} is the propagation of the first two moments used first for Fast Dropout \cite{wang2013fast}, where one samples from the layer outputs (implicitly approximated as moment-matched normal distributions by leveraging the Central Limit Theorem (CLT)), instead of the layer weights, and later in the context of BNNs \cite{kingma2015variational,roth2016variational,shridhar2018uncertainty,haussmann2019deep}. Propagation of more moments has been shown to be beneficial, \eg in resisting adversarial attacks, but also requires sampling for cubature \cite{wang2020bayesian}, or unscented \cite{dera2021premium} and particle \cite{carannante2021enhanced} filtering. That being said, many recent \textit{moment-propagating} BNN studies constitute Bayesian treatments of DL models with simple mechanisms, like dense \cite{wu2018deterministic, haussmann2019deep,carannante2021enhanced} and convolutional \cite{shridhar2018uncertainty,dera2021premium} layers interweaved by nonlinear activation functions \cite{roth2016variational,wu2018deterministic,haussmann2019deep,dera2021premium} -- even in non-Bayesian variance propagation \cite{tzelepis2021uncertainty}. Whereas ResNet-like skip connections on a dense layer based network have been proposed in \cite{wu2018deterministic}, less consideration has been given on doing the same for more advanced concepts like convolutional ResNets, SE, and attention. Furthermore, there does neither exist an explicit utilisation of sample-free predictive uncertainty as a signal for data-specific regularisation, nor (to our knowledge) a study on the calibration properties of such moment-propagating BNNs. Finally, we are only aware of MC-based Bayesian approaches for bioacoustics \cite{kiskin2021automatic}, a domain where it has been repeatedly suggested that a probabilistic \cite{kitzes2019necessity} or calibration-focused \cite{stowell2019automatic,stowell2022computational} framework should be considered, along with traditional accuracy-based performance evaluation.
% , not to mention the surprisingly challenging notion of max-pooling \textit{distributional} features.

% MNIST reference.

The contributions we make in this article are summarised as follows:

\begin{itemize}
    \item We test the first sample-free, uncertainty propagating, variational Bayesian treatment of a complex DL architecture that has excelled in the bioacoustics domain \cite{rizos2021multi}, by propagating random variable expectations and variances through squeeze-and-excitation blocks, and two types of local pooling: classic max-pooling \cite{weng1993learning} and a variant of a recently proposed attention-based pooling \cite{gao2019lip}.  To our knowledge, this is the first time a moment propagating version of the squeeze-and-excitation mechanism is proposed and evaluated, although MC-based Bayesian methods have done so before \cite{krokos2022bayesian}.
    \item We propose a sample-free regularisation method that explicitly uses the propagated epistemic uncertainty of our BNN model as an adaptive signal for label smoothing that is specific to each data sample, and, on most cases, achieves higher predictive and calibration performance compared to the aforementioned variational version without epistemic label smoothing. We showcase the value of our uncertainty-aware label smoothing by further testing it against a variant that is data-sample agnostic. We verify that deterministic, moment propagating BNNs -- including our proposed method -- exhibit high calibration quality.
    \item Our methodology is exemplified 
    %BS: is "at" the correct choice? "in"?
    in the bioacoustics domain, and evaluated specifically on challenging, real-world, so-called ``in-the-wild'' datasets, as literally is the case in bioacoustics for wildlife PAM, where the recordings contain multiple background sounds, other than the target. We use a spider monkey call detection dataset previously used in \cite{rizos2021multi}, and further introduce another dataset with annotations for 30 distinct species (29 bird species, and Bornean gibbons).
\end{itemize}

\section{Results}
\label{sec:results}

We focus on binary classification as the common type of task between our two animal call detection datasets. The \textit{Osa Peninsula Spider Monkey Whinny} (\textbf{OSA-SMW}) dataset was first introduced and described in \cite{rizos2021multi}, and a single binary call detection task is defined on it, where the focus is specifically the `whinny' call of Geoffroy's spider monkey (Ateles geoffroyi). We first introduce here the \textit{SAFE Project} \cite{ewers2011large} \textit{Multi-Species Multi-Task} (\textbf{SAFE-MSMT}) dataset, of which the description, and preprocessing details can be found in Supplemental Experimental Procedures -- Sub-Section SAFE-MSMT
Dataset. We consider the detection of calls for each species identified within the dataset as a separate binary task, and have identified thirty species such that, for all tasks, there are no empty negative or positive classes in any of the training, development, and testing sets. It is possible that there are zero, one, or more species' calls audible per audio clip, which constitutes a multi-label classification problem, approached via the multi-task framework, where each task is binary classification.

For evaluating our experiments, we opted to report the Area Under the precision-Recall (non-interpolated \textbf{AU-PR}) curve of the positive class, and the Area Under the Receiver Operating Characteristic (\textbf{AU-ROC}) curve as prediction performance measures that average over all possible probability thresholds for classification. Test performance is measured using the model that achieved best validation performance according to AU-PR, which is a stricter measure in class-imbalanced cases where the positive class is a minority, as AU-ROC is known to inflate due to the abundance of true negatives. We also report the unweighted average of F1 of the positive and negative classes at a probability threshold of $0.5$ (\textbf{F1}), as well as the Expected Calibration Error (\textbf{ECE}) for measuring calibration quality, as suggested by \cite{guo2017calibration}, with ten probability buckets. In order to provide a summary performance profile for the 30-task \textit{SAFE-MSMT} dataset, we report here the weighted average of the per-task performance measures, 
% (expanded in Supplementary Materials~\ref{sup:perspeciesperformanceinthemalaysoiamultispeciesmultitaskdataset})
where each weight is proportional to the number of positive instances per task. Even so, this is a quite austere evaluation as for some tasks there are only a handful of positive samples, which heavily restricts the predictive potential of supervised learning based approaches. We summarise in Table~\ref{tab:malaysiaresults} the predictive and calibration performance measure results that arose from our comparative analysis on animal call detection. In all cases, we performed 10 trials for
which we report mean and standard deviation.

\begin{table}[t]
\begin{center}
\begin{minipage}{0.9\linewidth}
\caption{Sample-free, variational Bayesian deep learning and uncertainty-aware label smoothing improves predictive and calibration performance in most cases.}
\label{tab:malaysiaresults}%
\begin{tabular}{@{}llllll@{}}
\multicolumn{6}{c}{SAFE Project Multi-Species Multi-Task}\\
\toprule
& SE-ResNet & W-AU-PR $\uparrow$  & W-AU-ROC $\uparrow$ & W-F1 $\uparrow$ & W-ECE $\downarrow$\\
\midrule
\parbox[t]{2mm}{\multirow{4}{*}{\rotatebox[origin=c]{90}{max-pool}}} & base    & $20.51 \pm 3.17$   & $77.88 \pm 2.02$  & $42.73 \pm 11.78$ & $26.86 \pm 10.59$  \\
& variational    & $19.46 \pm 2.30$   & $77.64 \pm 2.89$  & $\textbf{45.67} \pm 11.26$ & $\textbf{22.40} \pm 11.78$  \\
& smooth    &  $19.72 \pm 1.31$  & $\textbf{79.31} \pm 5.12$  & $33.16 \pm 18.60$ & $35.36 \pm 12.40$  \\
& ua-smooth    &  $\textbf{20.79} \pm 1.47$  & $77.44 \pm 2.78$  & $\textbf{45.68} \pm 7.02$ & $25.51 \pm 11.10$  \\
\midrule
\parbox[t]{2mm}{\multirow{4}{*}{\rotatebox[origin=c]{90}{att-pool}}} & base    &  $17.39 \pm 3.65$  & $75.11 \pm 2.84$  & $41.54 \pm 10.70$ & $28.48 \pm 7.68$  \\
& variational    & $17.42 \pm 2.80$   & $71.64 \pm 4.59$  & $39.70 \pm 11.40$ & $34.19 \pm 8.72$  \\
& smooth    &  $19.16 \pm 5.13$  & $72.14 \pm 4.84$  & $38.06 \pm 16.66$ & $25.93 \pm 11.94$  \\
& ua-smooth    &  $\textbf{20.64} \pm 2.64$  & $\textbf{77.64} \pm 1.70$  & $\textbf{52.60} \pm 3.27$ & $\textbf{19.92} \pm 6.56$  \\
\botrule
\end{tabular}

% \footnotetext{Source: This is an example of table footnote. This is an example of table footnote.}
% \footnotetext[1]{Example for a first table footnote. This is an example of table footnote.}
% \footnotetext[2]{Example for a second table footnote. This is an example of table footnote.}
\end{minipage}\qquad
\begin{minipage}{0.9\linewidth}
% \caption{Results on the \textit{Malaysia Multi-Species Multi-Task} dataset: Sample-free, variational Bayesian deep learning and uncertainty-aware local pooling and label smoothing improves predictions and calibration.}
\label{tab:smoothingresults}%
\begin{tabular}{@{}llllll@{}}
\multicolumn{6}{c}{Osa Peninsula Spider Monkey Whinny}\\
\toprule
& SE-ResNet & AU-PR $\uparrow$  & AU-ROC $\uparrow$ & F1 $\uparrow$ & ECE $\downarrow$\\
\midrule
\parbox[t]{2mm}{\multirow{4}{*}{\rotatebox[origin=c]{90}{max-pool}}} & base    & $61.66 \pm 11.47$   & $90.08 \pm 5.51$  & $64.48 \pm 17.67$ & $14.70 \pm 17.26$  \\
& variational    & $65.73 \pm 5.19$   & $90.77 \pm 3.77$  & $58.67 \pm 18.65$ & $19.35 \pm 13.57$  \\
& smooth    &  $54.69 \pm 14.76$  &  $87.69 \pm 6.05$  & $57.37 \pm 19.69$ & $20.92 \pm 22.18$  \\
& ua-smooth    &  $\textbf{66.59} \pm 6.42$  & $\textbf{92.01} \pm 3.68$  & $\textbf{66.03} \pm 19.17$ & $\textbf{14.51} \pm 17.63$  \\
\midrule
\parbox[t]{2mm}{\multirow{4}{*}{\rotatebox[origin=c]{90}{att-pool}}}
& base    & $66.12 \pm 7.19$   & $91.59 \pm 2.03$  & $73.79 \pm 5.47$ & $6.13 \pm 2.37$  \\
& variational    & $\textbf{66.85} \pm 6.62$   & $\textbf{91.92} \pm 1.90$  &$\textbf{75.85} \pm 5.93$ & $4.99 \pm 2.49$  \\
& smooth    &  $60.25 \pm 1.82$  &  $89.69 \pm 2.58$ & $72.40 \pm 4.16$ & $4.86 \pm 0.74$  \\
& ua-smooth    &  $66.52 \pm 5.21$  &  $\textbf{91.90} \pm 1.91$ & $75.05 \pm 3.80$ & $\textbf{4.63} \pm 1.59$  \\
\botrule
\end{tabular}
% \footnotetext{Source: This is an example of table footnote. This is an example of table footnote.}
% \footnotetext[1]{Example for a first table footnote. This is an example of table footnote.}
% \footnotetext[2]{Example for a second table footnote. This is an example of table footnote.}
\end{minipage}
\end{center}
\end{table}

\begin{figure}[t]%
\centering
\includegraphics[width=1.0\textwidth]{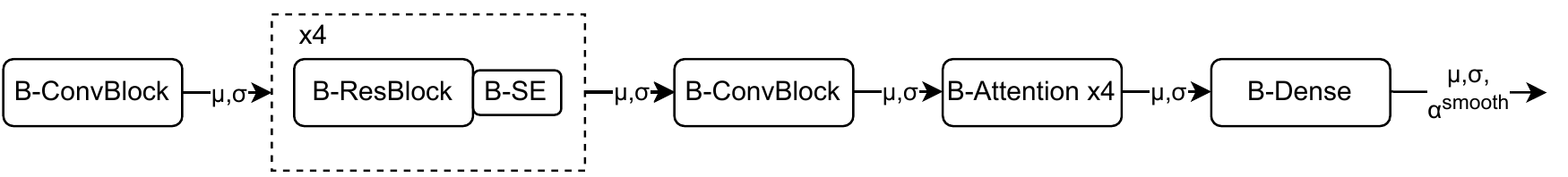}
\caption{This is an abstraction of the sample-free, moment propagating variational Bayesian ResNet model with squeeze-and-excitation and multi-head attention (B-SE-ResNet) we use as a basis throughout this study. The point-estimate version follows the same architecture, but each layer, block, and nonlinearity does not use variational learning for inference or make affordances for propagating uncertainty.}
\label{fig:coremodel}
\end{figure}

As the baseline in our comparisons, we used a variation of a modern, complex DL model that has performed well on the bioacoustics domain \cite{rizos2021multi}, which combines a ResNet architecture, SE blocks, and multi-head attentive pooling of sequential embeddings, and an output dense layer per binary task, for a total depth of 21 layers; hence, \textbf{base SE-ResNet}. A pictorial overview of the proposed Bayesian version of the model can be seen in Figure~\ref{fig:coremodel}, and we consider two variants of it, one with max-pooling (\textbf{max-pool}), and the other with attentive pooling (\textbf{att-pool}), \ie a pooling process that employs an additional dense layer that learns a weighted average of the activations to be pooled, an instance of the local pooling approach recently proposed in \cite{gao2019lip}. We compare their performances with those of the corresponding:
\begin{inparaenum}[a)]
  \item uncertainty propagating, variational Bayesian versions developed for this article in Sub-Section~\ref{subsec:craftingacompetitivebayesianseresnetbaseline},
  \item variants with the addition of our sample-free, uncertainty-aware label smoothing technique in Sub-Section~\ref{subsec:benefitsofuncertaintyusageinlabelsmoothing}.
\end{inparaenum} More details on the implementation of core mechanisms, the considerations made towards a Bayesian treatment, and the novelties proposed here can be found in Section~\ref{sec:experimentalprocedures}, and full technical details on the Supplemental Experimental Materials.

\subsection{Crafting a Competitive Bayesian SE-ResNet Baseline}
\label{subsec:craftingacompetitivebayesianseresnetbaseline}

As a first step towards a more uncertainty-aware approach, we modify \textit{base SE-ResNet} such that it becomes a variational Bayesian, uncertainty propagating version of itself. Linear operators like dense and convolutional neural layers are replaced with locally reparameterised versions, as described respectively in \cite{kingma2015variational} and \cite{shridhar2018uncertainty}, where the first two moments of the outputs are given in closed form, and are linearly dependent on the, also stochastic, respective layer inputs and weights, yet independent among themselves. The stochastic layer outputs are transformed by nonlinear activation functions like ReLU and sigmoid, where the first two moments of the activations are approximated as previously described in \cite{wang2013fast,roth2016variational,haussmann2019deep,tzelepis2021uncertainty}. Regarding max-pooling of such normally distributed variables, the authors of the studies performed in \cite{haussmann2020sampling,dera2021premium} independently proposed \textit{co-pooling} of the two moments, \ie propagating only the moments of the random variable with the highest expected value. As for attention-pooling, the weighted sum of normally distributed variables is well-known, and we learn the probabilistic weights using attention. This way, information on the first two moments of all pooled variables is propagated.

In the results Table~\ref{tab:malaysiaresults}, the Bayesian, uncertainty propagating version of SE-ResNet with max co-pooling (\textbf{variational max-pool}) exhibits a slightly higher performance than \textit{base max-pool} in terms of AU-PR, and AU-ROC for the OSA-SMW dataset, and also the best W-F1 and W-ECE among the max-pool based methods for SAFE-MSMT.

As for the attention-pooling variant (\textbf{variational att-pool}), we observe a lower performance compared to the point-estimate baseline in all measures except for W-AU-PR for SAFE-MSMT, and an improvement in all measures for OSA-SMW.

\subsection{Benefits of Uncertainty Usage in Label Smoothing}
\label{subsec:benefitsofuncertaintyusageinlabelsmoothing}

Label smoothing \cite{szegedy2016rethinking} in loss calculation is the use of a label distribution that is an interpolation between the true distribution, as given by the annotators, and the uniform distribution; in the binary classification task, the latter corresponds to $0.5$ probability for both the negative and the positive classes:
\begin{equation}
y_{i,c}^{smooth} = \alpha y_{i,c}^{uniform} + (1 - \alpha) y_{i,c}^{true},
\end{equation}
where $y_{i,c}$ refers to the label probability that class $c$ is correct for data sample $i$, and $\alpha$ denotes the smoothing probability hyperparameter, \ie the degree to which we want the model to \textit{not} overexert in trying to learn to classify that particular sample as per the ground truth $y_{i,c}^{true}$.

Here, we propose the first solution for data-specific label smoothing that is dependent on the uncertainty propagated throughout a BNN model, and is also \textit{MC sample-free}. A description of the means by which we define such an uncertainty-aware smoothing probability $\alpha_{i}$ for sample $i$, is found in Section~\ref{sec:experimentalprocedures}, and a schematic overview is depicted in the Graphical Abstract.

As seen in the results Table~\ref{tab:malaysiaresults}, our uncertainty-aware label smoothing method (\textbf{ua-smooth}) used on the BNN described in the previous Sub-Section, outperforms the (corresponding to pooling type) \textit{variational} method in terms of all measures, in most cases; \eg except for W-AU-ROC and W-ECE for the max-pool case on the SAFE-MSMT dataset, and AU-PR and F1 for att-pool on the OSA-SMW dataset. Even in those cases, it achieves better performance compared to the baseline in all cases. For the SAFE-MSMT case, the drop in W-AU-PR when using max-pool and the drop in W-AU-ROC, W-F1, and W-ECE when using att-pool are not only eliminated, but we see now the variational approach leveraging its untapped potential by this direct usage of the predictive uncertainty.

\subsubsection{Smoothing Should be Specific to Data-Samples}
\label{subsubsec:smoothingshouldbespecifictodatasamples}

How can we be sure, then, that the propagated model uncertainty contains information about which samples should use higher smoothing probabilities, and that it's not simply a case of label smoothing being beneficial in general?

To answer this question, we perform one more series of experiments, with a label smoothing variant (hence, \textbf{smooth}) that keeps the smoothing probability fixed across the training batch. Specifically, we calculate for every batch the average of the uncertainty-aware smoothing probabilities as per our proposed \textit{ua-smooth} method and apply that to all batch samples instead. This is not a hyperparameter-based, fixed-value label smoothing, as is commonly used, since it benefits from the uncertainty quantification provided by the BNN, the values of which change per training step as the model learns to model the training data, and it tracks the average value of the uncertainty-aware smoothing probability, thus allowing for a stricter comparison with the \textit{ua-smooth} method, which we propose as the best means of performing uncertainty-aware label smoothing using a BNN.

We observe that, for \textit{most cases}, our proposed \textit{ua-smooth} method outperforms this, more naive, \textit{smooth} variant that resembles fixed-parameter label smoothing; the exceptions are the W-AU-ROC score in SAFE-MSMT. Moreover, in many cases we observe that \textit{smooth} performs worse than the non-smoothing \textit{variational} version and even than the point-estimate baseline, both in predictive and calibration performance measures.

\section{Discussion}
\label{sec:discussion}

We now discuss:
\begin{inparaenum}[a)]
  \item the insights extracted from our experiments regarding our proposed methodology in Sub-Sections~\ref{subsec:propagateduncertaintyshouldbeexplicitlyused}-\ref{subsec:calibrationperformanceofbnnoutputs},
  \item relations to similar methods and means by which our method should engender a re-evaluation thereof in Sub-Section~\ref{subsec:rethinkingsimilarmethods}, and
  \item potential extensions, criticisms, and opportunities in Sub-Sections~\ref{subsec:maxpoolingnormalrandomvariables}-\ref{subsection:shouldwefocusontheeasydatathen}.
\end{inparaenum}

\subsection{Propagated Uncertainty Should be Explicitly Used}
\label{subsec:propagateduncertaintyshouldbeexplicitlyused}

Propagated predictive uncertainty, as per our \textit{variational} variant of SE-ResNet, affects loss value calculation in that it describes a predictive distribution from which multiple prediction instances can be sampled. This leads to an expected loss value calculation that is based on softer, less overconfident prediction outputs compared to a loss value based on point-estimate predictions; the utilisation of epistemic uncertainty involving all potential output samples has been cited as a major regularising strength of BNNs \cite{wilson2020case}.

In addition to the point-estimate \textit{base}, we have designed the sample-free \textit{variational} method to be a more advanced baseline, to more strictly compete with our proposed uncertainty based label smoothing method.

That being said, by means of an insight from \textit{our} experiments with the moment propagating `flavour' of BNNs, \ie the \textit{variational} Bayesian SE-ResNet, we observed promising (\eg overall improvement on the OSA-SMW dataset), yet inconsistent results.
As such, we recommend that the Bayesian property, on its own, should be considered to be a type of hyperparameter, not to be employed agnostically, but only after experimental validation on the task under examination, including consideration of the relevant performance measures thereof.

However, the Bayesian formulations offer us another highly informative signal, something exclusive to them and unavailable to the baseline: the value itself of \textit{predictive variance, \ie a proxy of epistemic uncertainty}. There is a more explicit manner of utilising it, which can, and indeed should, be used in the loss calculation, as, in our experiments, the \textit{ua-smooth} method performs better than the corresponding \textit{variational} in almost all cases and performance measures.

Usually, predictive uncertainty is used in downstream tasks, \eg as a signal for data acquisition in active learning \cite{gal2017deep,haussmann2019deep}, or towards the design of uncertainty-aware (\eg risk-averse) reinforcement learning agents \cite{depeweg2018decomposition}. Inversely, we believe that uncertainty should be used as a signal that guides learning in the self-same task, and by the self-same model that is undergoing training; as per our experiments, not doing so is missing the opportunity given by the usage of a BNN, and is also disregarding one half of the BNN output. The sample-free manner of uncertainty, offers a more elegant, and less parameter-costly means of doing so compared to MC based methods.

More than that, our experiments with the batch-wide fixed smoothing method (\textit{smooth}) indicate that a higher degree of label smoothing is beneficial to data samples for which the BNN is less confident in modelling, and that, thus, the \textit{ua-smooth} variant is preferrable. That being said, for the SAFE-MSMT dataset, it seems that \textit{smooth} performs better than the baseline with point-estimate weights, indicating that Bayesian regularisation is beneficial for that dataset in any shape or form, most probably due to the positive class sample scarcity in all binary classification tasks of this dataset.

\subsection{Sample-Free BNN Outputs are Calibrated}
\label{subsec:calibrationperformanceofbnnoutputs}

A recommendation on which Bayesian approach to use agnostically is not easy to make, although in both the OSA-SMW and the SAFE-MSMT datasets, we observe that the calibration performance of the point-estimate baselines are overall worse compared to the variational versions. In three out of four cases, our proposed \textit{ua-smooth} method achieves better ECE performance than naive \textit{smooth}, although the regular \textit{variational} method achieves the highest calibration in one out of four cases.

\subsection{Rethinking Label Smoothing}
\label{subsec:rethinkingsimilarmethods}

Whereas label smoothing with a fixed, pre-defined smoothing probability has been originally proposed as a regularisation measure to improve predictive performance \cite{szegedy2016rethinking}, its success \cite{lukasik2020does,wei2021smooth} has been inconsistent, as it has in other cases been shown to deteriorate it \cite{wang2021rethinking,singh2021dark}, without necessarily improving calibration \cite{singh2021dark}.

That being said, label smoothing has been considered as one of the reasons for the high performance achieved by the student model in knowledge distillation \cite{hinton2015distilling}, as the latter utilises the smooth prediction probabilities output by the teacher model in place of ground truth labels. These output distributions are smoother, \ie closer to the uniform, for data samples that the teacher model finds difficult to model, thus constituting \textit{data-specific smoothing}. Moving away from the two-step, teacher-student framework (that is focused on model compression), in this study, we have shown the usefulness of a means for smoothing that requires no more that a single model, a single training process, and is also MC sample-free. As indicated by our experiments, we believe that the underlying conception of label smoothing is still promising, with the caveat that they need to be made in an adaptive, intelligent, and data-specific manner; \textit{a fortiori} in the uncertainty propagating BNN context, where a guiding signal is provided by design.

The study that is closer conceptually to our own is the one performed in \cite{seo2019learning}, in which the authors use the MC-based BNN approach proposed in \cite{gal2016dropout} called MC-dropout, and calculate the loss as an interpolation of the cross entropy between the predictions and the true labels, and the cross entropy between the predictions and the uniform distribution, where these two factors are weighed based on a value that is a normalisation of the MC-based estimate of the variance. Even though their loss calculation uses the predictions of a single execution, it also requires $K$ executions for estimating the variance. As such, the authors use $5$ MC samples and, subsequently, $5$ propagations of the input through the entire model during training. Instead, we use both the expectation and the variance of the outputs in our loss calculation, as propagated through the entire network in closed form approximation, constituting a more deterministic and elegant solution. These differences -- firstly, regarding the BNN approach, given the long-standing criticisms of MC-dropout on whether its assumptions and approximations truly constitute a Bayesian method \cite{osband2016risk,verdoja2020notes,folgoc2021mc}, though mostly due to the prohibitive additional computational and storage load introduced by MC sampling -- have led us to not consider a direct comparison with this method necessary.

\subsection{Locally Pooling Normal Random Variables}
\label{subsec:maxpoolingnormalrandomvariables}

Apart from max-pooling, we have also showcased the efficacy of a BNN approach on a newer, more elaborate local pooling method \cite{gao2019lip}. Whereas on the SAFE-MSMT dataset we observe an improvement in using attention over maximum pooling only in the W-AU-ROC, W-F1, and W-ECE measures for the \textit{ua-smooth} method, on OSA-SMW attention pooling is the clear winner. Most importantly, our experiments with attention pooling render our comparisons highly robust, as our proposed \textit{ua-smooth} method exhibits at least competitive performance on an additional dimension in architecture design.

\subsection{Should we Focus on the Easy Data Then?}
\label{subsection:shouldwefocusontheeasydatathen}

The underlying philosophy of our uncertainty-aware smoothing method is that high predictive uncertainty implies a training data sample that is, for whatever reason (\eg difficulty, subjectivity, scarcity etc.), difficult to model, and, as such, that our BNN should not over-penalise itself trying to memorise it. Similar assumptions have been made by past studies that focus on aleatory uncertainty \cite{kendall2017uncertainties}, and soft labels due to rater disagreement \cite{chou2019every}, or label smoothing \cite{szegedy2016rethinking,seo2019learning}. That being said, there has also been an alternate way of thinking, as in: data samples that are \textit{too easy} to model, should be the ones either ignored or downweighed, such that we avoid a flood of common samples dominating the loss calculation. A method that follows this paradigm is the focal loss \cite{lin2017focal}, of which newer versions are also heteroscedastic, \ie dependent on the input, as the degree of focus is itself dependent on an auxiliary output of the model \cite{li2021generalized}. This is similar to our approach, albeit we are not using a separate output ``head'', but leverage the Bayesian predictive uncertainty. A combination of these two philosophies, and a means by which we can learn the degrees to which we should downweigh both the easy, as well as the difficult samples side-by-side is something we would like to focus in a future extension of this study, potentially by incorporating uncertainty decomposition methods \cite{kendall2017uncertainties}.

\subsection{Conclusion \& Future Work}
\label{subsec:conclusion}

Whereas the predictive uncertainty signal calculated by Bayesian neural networks is often used to make decisions in a downstream task, like identifying samples to annotate in active learning, or addressing risk in reinforcement learning, in this article, we have used it to guide learning in the self-same task the neural network is being trained on. To that end, we have focused on deterministic (\ie non Monte Carlo based) Bayesian Neural Networks that propagate feature variances along with expectations, and utilised the end-to-end propagated output uncertainty to inform the degree of label smoothing that is applied in a data-specific manner. Both our novel sample-free variational Bayesian squeeze-and-excitation ResNet, and our recommended variant with uncertainty-aware label smoothing have yielded a noticeable improvement over the point-estimate baseline, and we submit that the incorporation of the sample-free uncertainty value that is available to such networks by design should be incorporated in the loss explicitly, as we have done, as a standard.

Our methodology has been evaluated on two animal call detection bioacoustics datasets, one of them introduced here for the first time, as well as in two variations pertaining to local latent unit pooling, which, we believe, increases the confidence of the insights extracted in our study. This work both advances work on moment propagating Bayesian neural networks that are of great use in the domain of deep learning, and is of special interest to the application field of bioacoustics, where low signal-to-noise ratio data often also receive weak annotation, leading to a need for soft, modest predictions that are highly calibrated; noted so far to be missing \cite{kitzes2019necessity,stowell2019automatic,stowell2022computational}. Well-calibrated model outputs, with meaningful prediction probabilities are required for downstream processing either by automatic decision making software, or human experts, especially in a collaborative human-AI setting, such as active learning. Though other types of Bayesian neural networks are known to perform well in terms of calibration \cite{maddox2019simple,osawa2019practical}, we have shown here that this also holds for the moment-propagating variety, with and without the use of our intelligent label smoothing.

We believe this study can stimulate research in uncertainty-aware local pooling and attention methods, in identifying informative data samples \cite{rizos2019modelling} in an integrated manner with focal loss \cite{li2021generalized},  and in trustworthy decision making in bioacoustics. Finally, we believe it is of interest to approach the newly-introduced SAFE-MSMT dataset via a few-shot learning framework, to extract as much information as possible from the limited size labelling.

\section{Experimental Procedures}
\label{sec:experimentalprocedures}

\subsection{Resource Availability}
\label{subsec:resourceavailability}

\subsubsection{Lead Contact}
Further information regarding the computational methodology and use of codebase should be directed to and will be fulfilled by the lead contact, G.R. ( georgios.rizos12@imperial.ac.uk ). Information regarding the SAFE-MSMT dataset should be addressed to R.E ( r.ewers@imperial.ac.uk ), and regarding OSA-SMW to J.L ( j.lawson17@imperial.ac.uk ).

\subsubsection{Materials Availability}
This study did not generate new unique materials or reagents.

\subsubsection{Data and Code Availability}

All original code has been deposited at \url{https://github.com/glam-imperial/sample-free-uncertainty-label-smoothing} under DOI \url{https://zenodo.org/badge/latestdoi/519486336} and is publicly available as of the date of publication. The SAFE-MSMT data reported in this paper will be shared by R.E. upon request (see \url{https://safedata-validator.readthedocs.io/en/latest/}). The OSA-SMW data reported in this paper will be shared by J.L upon request.

\subsection{Epistemic Uncertainty-Aware Label Smoothing}
\label{subsec:epistemicuncertaintyawarelabelsmoothing}

We need to quantify the belief that an input sample has been noisily annotated, and as such the prediction error for it should contribute less to the loss value calculation; we design such a measure by adhering to the following desiderata:
\begin{inparaenum}[a)]
    \item it is in the $[0, 1]$ range, such that it can serve as the label smoothing probability,
    \item it is positively correlated to the propagated, predictive variance in order to reflect BNN uncertainty about the input sample categorisation, and
    \item it is also positively correlated to overconfident (\ie close to $1$) predictions, such that moderate predictions do not receive feedback reinforcement.
\end{inparaenum}

Consider the expected logit output $\mathbb{E}[h^{t}_{i,L}]$ of a dense prediction layer, where $L$ denotes the last layer index and $t$ the task corresponding to that prediction layer. If we \textit{do} utilise the logit variance $\mathbb{V}[h^{t}_{i,L}]$ and transform the normally distributed random variable via a sigmoid function (as detailed in the Supplemental Experimental Procedures -- Sub-Section Sample-Free Variational Attentive SE-ResNet), we get the fully propagated, Bayesian expectation and variance of $y_{i,\text{POS}}^{\text{t,Bayes}}$, \ie the probability that the input sample is from the positive (POS) class. Inversely, if we opt for a Maximum A Posteriori (MAP) approach for that final layer, by \textit{not} utilising the logit variance, we transform the logit expectation via the sigmoid and denote the probability by $y_{i,\text{POS}}^{\text{t,MAP}}$.

$y_{i,\text{POS}}^{\text{t,MAP}}$ would still benefit from the moment propagation up until the final layer in terms of the learnt features and logits $h_{l}$ (for $l$ up to, but excluding $L$), as well as from the Bayesian regularisation for all layers. However, final layer MAP makes the information encoded in the propagated uncertainty unavailable in the calculation of the predictive probability distribution. Inversely, $y_{i,\text{POS}}^{\text{t,Bayes}}$ gets the full benefits of the Bayesian approach. A fully Bayesian treatment of even just the final layer has been shown to have a positive benefit on addressing overconfidence, even when the rest of the model is parameterised with point estimate weights \cite{kristiadi2020being}.

We, thus, attempt to capture this additional, Bayesian uncertainty information by defining the data sample-specific smoothing probability as:
\begin{equation}
\alpha_{i}^{t} = \vert y_{i,\text{POS}}^{\text{t,MAP}} - y_{i,\text{POS}}^{\text{t,Bayes}} \vert.
\label{eq:manhattandistance}
\end{equation}

\begin{figure}[t]%
\centering
\includegraphics[width=0.7\textwidth]{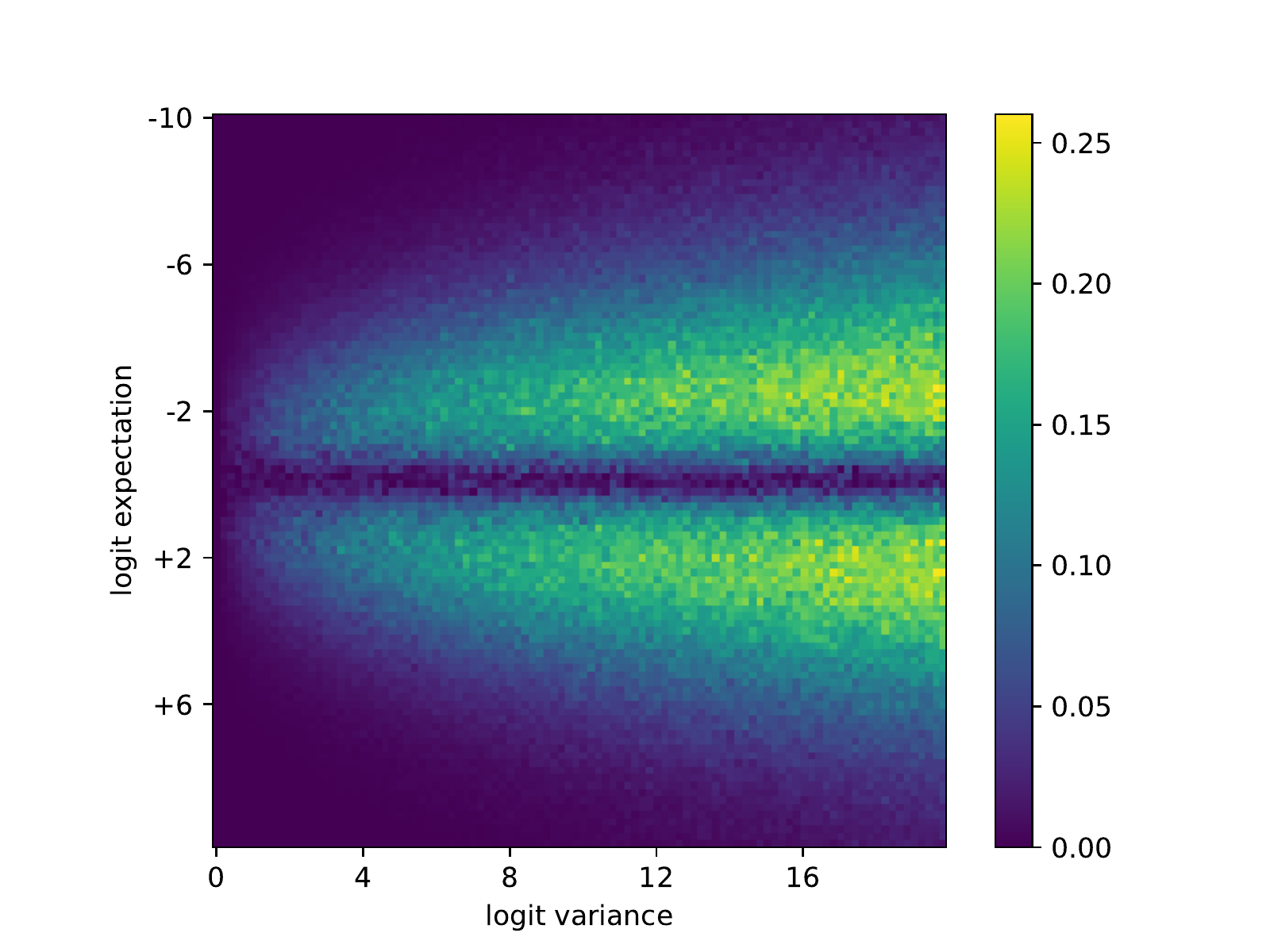}
\caption{The value of our proposed adaptive, uncertainty-aware smoothing probability given the expectation and variance of the logit. For close to $0$ logit uncertainties, the smoothing probability $\alpha_{i}^{t}$ is also close to $0$. For higher logit uncertainties $\mathbb{V}[h^{t}_{i,L}]$, $\alpha_{i}^{t}$ is higher for predictions that are closer to the extreme values of either $0$ or $1$. For moderate predictions close to $.5$, $\alpha_{i}^{t}$ is closer to $0$, thus encouraging learning from the true signal, instead of reinforcing a moderate prediction behaviour.}
\label{fig:manhattan_smoothing_probability}
\end{figure}

For a binary call detection task, this is equivalent to the Manhattan distance between the corresponding two-element discrete predictive probability distributions, multiplied by two. A visualisation of our adaptive smoothing probability given ranges of logit expectations and variances can be found in Figure~\ref{fig:manhattan_smoothing_probability}.

\section*{Acknowledgments}
\label{sec:acknowledgments}

G.R. would like to acknowledge the Engineering and Physical Sciences Research Council (EPSRC) Grant No.\ 2021037.
%BS: don't forget this :)

\subsection*{Author Contributions}

G.R. conceived and designed the proposed methodology, coded the moment-propagating Bayesian neural networks and smoothing methods, executed the experiments, wrote the article, and prepared figures. J.L. collected and annotated the OSA-SMW dataset, contributed in designing the predictive task, and writing related parts. S.M. annotated the SAFE-MSMT dataset. P.S. contributed in coding the dense Bayesian neural layer, the ARD prior, and proposed the use of cold posteriors for training. X.W. preprocessed the SAFE-MSMT dataset, prepared initial versions of related figures and descriptions of related parts, and executed exploratory experiments on SAFE-MSMT. C.B.L, R.E., and B.W.S. supervised the research. All authors discussed the results and commented-on/edited the manuscript.

\subsection*{Declaration of Interests}
The authors declare no competing interests.

% superscript
% \bibliographystyle{unsrt}
% \bibliographystyle{unsrtnat}


\pagenumbering{gobble}

% \begin{appendices}

\section*{Supplemental experimental procedures}
\label{sup:modelimplementationdetails}

\subsection*{Baseline Attentive SE-ResNet for Bioacoustics}
\label{subsec:baselineattentiveseresnetforbioacoustic}

Our baseline is an SE-ResNet with multiple-head attention that is a variant of the one proposed in \cite{rizos2021multi}; an architecture overview is depicted in Figure 1 in the main article, and further details with parameter values and tensor shapes is summarised in Table~\ref{tab:seresnet}. It is designed to process log-Mel spectrograms as input, \ie 2-dimensional audio representations. We can divide it in three modules:
\begin{inparaenum}[a)]
  \item the \textbf{core, audio processing module} that produces a sequence of learnt audio embeddings and is based on convolutional layers, residual blocks, local pooling (maximum or attentive), and squeeze-and-excitation blocks,
  \item the \textbf{multiple-head, attention mechanism} for weighted average pooling of the embeddings, and
  \item the \textbf{top module}, a set of dense layers that process the averaged, recording-wide neural representation, where each layer makes a prediction corresponding to a separate binary call detection task. There is one such layer for the OSA-SMW, and thirty for the SAFE-MSMT dataset.
\end{inparaenum}
We extract spectrograms from sound waveform sampled at a rate of $16$KHz, by using a Fast Fourier Transform window of $128$ms, sliding at a hop length of $10$ms. Given a three-second clip, we extract $128$ Mel coefficients and end up with a log-Mel spectrogram with sequence length equal to $300$.

\begin{table}[h]
\begin{center}
\begin{minipage}{1.0\linewidth}
\caption{SE-ResNet with multiple head attention implementation. The sample-free variational versions share the same architecture, albeit by propagating moments throughout.}
\label{tab:seresnet}%
\centering
\begin{tabular}{cc}
    Model Operation & Shape  \\
    \toprule
    Log-Mel Spectrogram & (300, 128)            \\
    \midrule
    (ConvBlock @ 64, ReLU) \& Pool & (150, 64, 64)  \\
    \midrule
    (SEBlock @ 64, ReLU)$\times$2 \& Pool & (75, 32, 64)  \\
    \midrule
    (SEBlock @ 128, ReLU)$\times$3 \& Pool & (37, 16, 128)  \\
    \midrule
    (SEBlock @ 256, ReLU)$\times$5 \& Pool & (18. 8, 256)  \\
    \midrule
    (SEBlock @ 512, ReLU)$\times$2 \& Pool & (9, 4, 512)  \\
    \midrule
    (ConvBlock @ 1024, ReLU) & (9, 4, 1024)  \\
    \midrule
    Reshape embedding & (9, 4096)  \\
    \midrule
    4-head attention-based pooling & (4096$\times$4,)  \\
    \midrule
    Dense layer per task & (1,) $\times$ tasks  \\
    \bottomrule
    
  \end{tabular}
\end{minipage}
\end{center}
\end{table}

As seen in Table~\ref{tab:seresnet}, the log-Mel spectrogram is first processed by a block (\textbf{ConvBlock}) of two convolutional layers, each with 64 filters and ReLU activations, and followed by a pooling operation without padding. The pooling operation can be either max- or attentive-pooling. Then, the hidden units are processed by four blocks (\textbf{SEBlock}), where each is comprised of two residual layers with squeeze-and-excitation mechanisms, and is followed by a pooling operation. The core module concludes with another ConvBlock, where the convolutions learn 1024 filters, but this time not followed by pooling. In all cases, the convolutional layers learn $3\times3$ filters and corresponding biases, and the pooling operations are subsampling at a $2\times2$ ratio.

The above module transforms a log-Mel spectrogram input into a hidden tensor with sequence length of $9$, width of $4$, and $1024$ features. We want to perform global pooling across the sequence length, and so first reshape the tensor to $(9, 4096)$. We then learn four weighted sequence averaging operations, using four attention heads. Each head corresponds to a learnt linear transformation of each embedding frame to a single energy value, and the calculation of a probability vector by passing the energy values from the sequence through a softmax function. These probabilities are used for weighted averaging, leading to an averaged embedding per attention head; those are then concatenated to provide a single, sequence-wide representation of the input audio clip. This is processed by the top module, where the dense layer that corresponds to each task avails of the common base model for shared feature extraction. Each dense layer produces one logit per data sample, which is passed through a sigmoid function such that we obtain the probability that the sample is positive.

In the following Sub-Section, we describe how all mechanisms in the above-mentioned attentive SE-ResNet are adapted such that they constitute corresponding variational Bayesian versions that propagate uncertainty.

\subsection*{Sample-Free Variational Attentive SE-ResNet}
\label{subsec:samplefreevariationalattentiveseresnet}

In Bayesian deep learning, each model parameter constitutes a random variable, instead of a single point estimate value, and we opt for the Automatic Relevance Determination (ARD) \cite{mackay1994bayesian,neal1996bayesian,titsias2014doubly,neklyudov2019variance} prior distribution $p(W)$ for the set of weights $W$. By deciding on a likelihood function, and following Bayes' rule, we can define the posterior weight distributions $p(W \vert X, Y)$ \cite{mackay1994bayesian,neal1996bayesian} after observing the data inputs $X$ and labels $Y$. Closed-form calculation for this high-dimensional and highly complex posterior is intractable, so we opt for approximate Bayesian computation, by learning a variational distribution $q_{\omega}(W)$ such that it is close to the true posterior \cite{titsias2014doubly,blundell2015weight}, \ie by minimising $D_{\text{KL}}[q_{\omega}(W) \Vert p(W \vert X, Y)]$, where $\omega$ denotes the set of the new, variational parameters. By utilising the prior and selecting a likelihood function $p(Y \vert W, X)$, the above KL divergence optimisation cost is rewritten as the Evidence Lower Bound (ELBO): $\mathbb{E_{W \sim \omega}}[p(Y \vert W, X)] - c D_{\text{KL}}[q_{\omega}(W) \Vert p(W]$, where $c$ is a regularisation parameter for realising a \textit{cold posterior}; \ie placing less importance on the regularising power of the prior, something that has been shown to be beneficial in BNNs \cite{wenzel2020good}.

As for the first term of the ELBO, whereas there have been proposed approximated versions of the binary \cite{tzelepis2021uncertainty} and categorical \cite{roth2016variational,wu2018deterministic} cross-entropy loss, we elect, instead, the alternative approximation of calculating the loss by sampling from the logit normal distributions, transforming separately through the sigmoid, and averaging the sample-based loss factors. Such logit output sampling is very lightweight, compared to MC sampling of the entire network, so we believe we still adhere to our sample-free, uncertainty propagating desideratum, and a similar rationale has previously been employed in \cite{kendall2017uncertainties} for Bayesian deep classification. In every other case, we perform moment propagation, as explained in the following.

We opt for a normal variational distribution $q_{\omega}(W) = \prod_{i,j,l} \mathcal{N}(w_{ijl} \vert \mu_{ijl}, \sigma_{ijl}^2)$, where the indices $i,j$ denote the weight array coordinates, and $l$, the layer index. The single-dimensional normal distributions defined by $i,j,l$, and from which weight instances can be sampled, are independent. In order to reduce the space and time computational load from such MC weight sampling, as well as reducing the gradient variance due to artifacts entailed by naively -- albeit more cost-effectively -- sampling weights once per batch instead of once per sample \cite{kingma2015variational}, we opt to also treat the neural layer outputs as random variables according to the fast \cite{wang2013fast} and variational dropout \cite{kingma2015variational} reparameterisation approaches. These output random variables are approximated via the CLT by normal distributions, from which samples can be taken; however, we opt for propagating the means and variances through the rest of the network. Thus, for a normally distributed input $h_{il-1}$, and Bayesian dense layer weights $w_{ijl}\sim\mathcal{N}(w_{ijl} \vert \mu_{ijl}, \sigma_{ijl}^2)$, the output neuron $a_{jl}$ can be sampled, as in \cite{kingma2015variational,roth2016variational,haussmann2019deep} from:

\begin{equation}
p(a_{jl}) = \mathcal{N}(a_{jl} \vert \phi_{jl}, \lambda^{2}_{jl}),
\label{eq:linearprop}
\end{equation}
where,
\begin{align}
\phi_{jl} &=  \sum_{i}\mu_{ijl}\mathbb{E}[h_{il-1}], \nonumber \\
\lambda^{2}_{jl} &= \sum_{i}\mathbb{V}[h_{il-1}]\mu_{ijl}^2 + \sigma_{ijl}^{2}\mathbb{E}[h_{il-1}^{2}],
\label{eq:linearprop2}
\end{align}
and similarly for Bayesian convolutional layers \cite{shridhar2018uncertainty,haussmann2019deep,neklyudov2019variance}, albeit following a sliding window. The addition of bias weights in dense and convolutional layers, as in our implementation, follows well-known addition of normal variables.

Parameterising stochastic weights by storing both means and variances is something that has not been shown \cite{neklyudov2019variance} to be worth the double space requirement, at least compared to a variant parameterisation $\sigma_{ijl}^{2}=\rho_{l}\mu_{ijl}^{2}$, where $\rho_{l}$ can be considered to be a positive, adaptive, layer-wise value (reminiscent of the variational dropout rate \cite{kingma2015variational}), and thus the variational parameters $\omega=[\mu_{ijl},\rho_{l}]$. The authors of \cite{schmitt2021sampling} have also supported the approach of not storing variance-related weight parameters.

We transform these pre-activations $a_{jl}$ via nonlinear activation functions, and with moment-matching of expectation and variance we estimate the normal activated variables $b_{jl} \sim \mathcal{N}(b_{jl} \vert \theta_{jl}, \xi^{2}_{jl})$. For ReLU, we follow \cite{roth2016variational}:

\begin{align}
\theta_{\text{ReLU},jl} &=  \frac{\phi_{jl}}{2}\left(1 + \text{erf}\left(\frac{\phi_{jl}}{\sqrt{\lambda_{jl}}}\right)\right) + \sqrt{\frac{\lambda_{jl}}{2 \pi}}\text{exp}\left(-\frac{\phi^{2}_{jl}}{2\lambda_{jl}}\right), \nonumber \\
\xi^{2}_{\text{ReLU},jl} &= \frac{\lambda_{jl} + \phi^{2}_{jl}}{2}\left(1 + \text{erf}\left(\frac{\phi_{jl}}{\sqrt{\lambda_{jl}}}\right)\right) + \phi_{jl}\sqrt{\frac{\lambda_{jl}}{2 \pi}}\text{exp}\left(-\frac{\phi^{2}_{jl}}{2\lambda_{jl}}\right),
\label{eq:relu}
\end{align}
where $\text{erf}(\cdot)$ is the Gauss error function. Regarding the sigmoid, approximations of the variance of the sigmoid activation has been provided by \cite{wang2013fast} and also by \cite{daunizeau2017semi}; the latter also adopted by \cite{tzelepis2021uncertainty}. We, however, use the former version due to its being more computationally lightweight. The approximation of the mean is common between these versions. Hence:
\begin{align}
\theta_{\text{sigmoid},jl} &\approx \text{sig}\left(\frac{\phi_{jl}}{\sqrt{1 + \pi \lambda^{2}_{jl}/8}}\right), \nonumber \\
\xi^{2}_{\text{sigmoid},jl} &\approx \mathbb{E}[\text{sig}\left(\alpha(\phi_{jl} - \beta)\right)] - \mathbb{E}[\text{sig}[\phi_{jl}]]^{2},
\label{eq:sigmoid}
\end{align}
where $\text{sig}(\cdot)$ is the regular sigmoid function, and $\alpha=4-2\sqrt{2}, \beta=-\text{log}(\sqrt{2} -1)$. It should be noted here, that both the activation expectation $\theta_{jl}$ and variance $\xi_{jl}$ are described as a nonlinear combination of both the pre-activation expectation $\phi_{jl}$ and variance $\lambda_{jl}$, thus ensuring that information encoded in both the pre-activation moments is propagated, \eg as the input to the next layer, \ie $\mathbb{E}[h_{il}], \mathbb{V}[h_{il}]=\theta_{jl},\xi^{2}_{jl}$.

\subsection*{Uncertainty Propagation and Attention}
\label{subsec:uncertaintypropagationandattention}

The SE-ResNet model further uses attention-like mechanisms in three instances: attentive weighted averaging which learns importance weights for aggregating feature vectors, used in
\begin{inparaenum}[a)]
  \item \textit{multi-head attention for global sequence pooling}, and
  \item in \textit{local attentive pooling} when used instead of max-pooling, as well as
  \item \textit{SE blocks}, that learn a reweighing of convolutional filter activations.
\end{inparaenum} We believe that the uncertainty of these filter vectors and sequential embeddings represents valuable information that should be propagated, and, if possible, utilised in the calculation of the importance weights.

\begin{figure}[t]%
\centering
\includegraphics[width=0.4\textwidth]{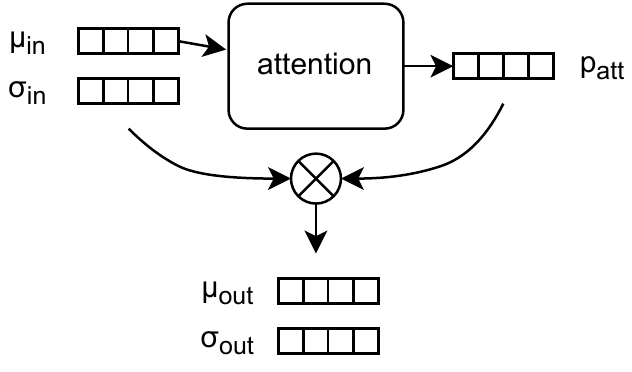}
\caption{Uncertainty propagating attention, used both for multiple attention head global pooling, and local attention pooling.}
\label{fig:uncertaintyawareattention}
\end{figure}

Regarding the first two cases, the attentive weighted average $a_{agg}$ of a set of feature vectors $b_{t}$ is performed as a linearly weighted sum of a set of normal variables, where the weights are probabilistic and sum to $1$:

\begin{align}
\mathbb{E}[a_{agg}] &= \sum_{t}{p_{att,t}\theta_{t}}, \nonumber \\
\mathbb{V}[a_{agg}] &= \sum_{t}{p_{att,t}^{2}\xi_{t}^{2}},
\label{eq:attention}
\end{align}
where $b_{jl} \sim \mathcal{N}(b_{jl} \vert \theta_{jl}, \xi^{2}_{jl})$ is a normally distributed activation vector as described in the Supplementary Sub-Section~\ref{subsec:samplefreevariationalattentiveseresnet}, and
\begin{align}
p_{att, t} &= \frac{\text{exp}(\varepsilon_{t})}{\sum_{t'}{\text{exp}(\varepsilon_{t'})}}, \nonumber\\
\varepsilon_{t} &= \theta_{t}A,
\label{eq:softmax}
\end{align}
and where $\text{exp}(\cdot)$ is the exponential function, and $A$ is a variational dense attention layer that transforms the activation vector into a single energy value $\varepsilon_{t}$.  In contrast to moment-propagating max-pooling as used in \cite{dera2021premium,haussmann2020sampling}, this type of local attention pooling achieves the propagation of uncertainty information from all pooled variables, and constitutes an uncertainty-aware extension of the local pooling method proposed in \cite{gao2019lip}. This process is sketched in Figure~\ref{fig:uncertaintyawareattention}.

\begin{figure}[t]%
\centering
\includegraphics[width=0.3\textwidth]{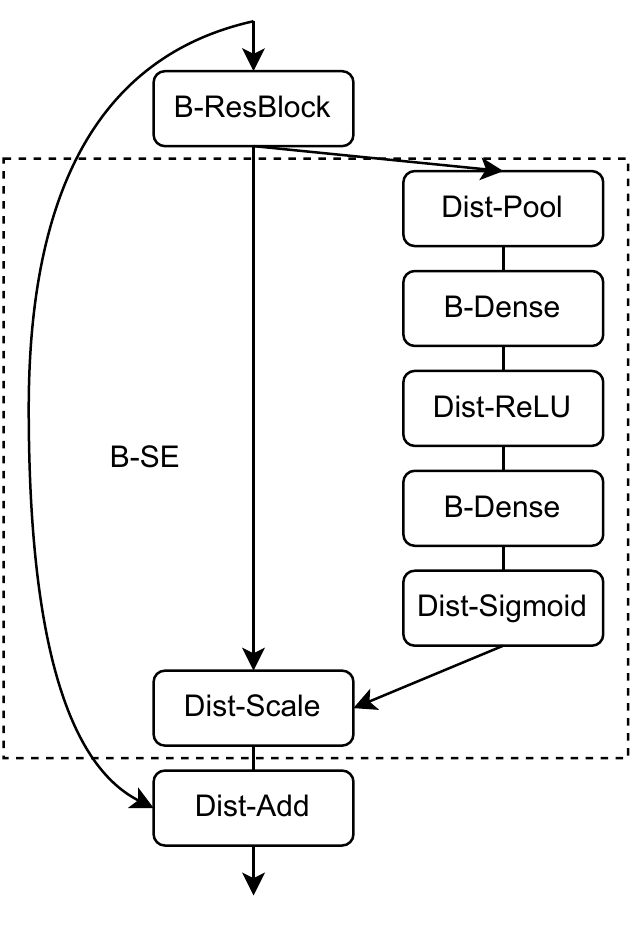}
\caption{A Bayesian variant of squeeze-and-excitation (B-SE) is achieved by replacing all blocks and layers with moment propagating Bayesian versions (B-ResBlock, B-Dense), and all nonlinearities and pooling-based sub-sampling with their distributional counterparts (Dist-ReLU, Dist-Sigmoid, Dist-Pool), as well as scaling and addition with the respective operations on normal distributions (Dist-Scale, Dist-Add). The specific SE base architecture variant we use was first proposed in \cite{naranjo2020acoustic}.}
\label{fig:uncertaintyseblock}
\end{figure}

As for the uncertainty propagating SE mechanism, a schema is depicted in Figure~\ref{fig:uncertaintyseblock}. We describe how weighted averaging, dense layers, and the ReLU and sigmoid activation functions propagate moments in the Supplementary Sub-Section~\ref{subsec:samplefreevariationalattentiveseresnet}. The only additional step to consider, here, is the final scaling of the input by multiplying it with the excitation importance weights after the sigmoid activations. The expectation and variance of the scaled residual neural filter variables $h_{se\text{-}res,jl}$ is the product of two normal random variables; and, hence, known:

\begin{align}
\mathbb{E}[h_{se\text{-}res,jl}] &= \mathbb{E}[h_{res,jl}]\mathbb{E}[\varepsilon_{jl}], \nonumber\\
\mathbb{V}[h_{se\text{-}res,jl}] &= \left(\mathbb{E}[h_{res,jl}] + \mathbb{V}[h_{res,jl}] \right) \left(\mathbb{E}[\varepsilon_{jl}] + \mathbb{V}[\varepsilon_{jl}] \right) - \mathbb{E}[h_{res,jl}]\mathbb{E}[\varepsilon_{jl}],
\label{eq:productofnormals}
\end{align}
where $h_{res,jl}$ is the residual neural filter variables output by a residual block, yet before scaling by the excitation $\varepsilon_{jl}$ as calculated by SE.

This way, learnt filter uncertainty is propagated after scaling, but, most importantly, the uncertainty itself is taken into account in the scaling value calculation.

\subsection*{Bioacoustic Data Collection}
\label{sup:bioacousticdatacollection}

A short summary of the datasets is found in Tables~\ref{tab:dataset1}-\ref{tab:dataset2}.

\begin{table}[t]
\begin{center}
\begin{minipage}{0.9\linewidth}
\caption{A summary of the SAFE-MSMT dataset. We refer to positive samples as clips that contain at least one call type.}
\label{tab:dataset1}%
\begin{tabular}{@{}llll@{}}
\multicolumn{4}{c}{SAFE Project Multi-Species Multi-Task}\\
\toprule
Dataset & train  & devel & test\\
\midrule
positives    & $383$   & $208$  & $218$  \\
\midrule
negatives    & $1,160$   & $556$  & $541$  \\
\botrule
\end{tabular}
\end{minipage}
\begin{minipage}{0.9\linewidth}
\caption{A summary of the OSA-SMW dataset.}
\label{tab:dataset2}%
\begin{tabular}{@{}llll@{}}
\multicolumn{4}{c}{Osa Peninsula Spider Monkey Whinny}\\
\toprule
Dataset & train  & devel & test\\
\midrule
positives    & $314$   & $125$  & $152$  \\
\midrule
negatives    & $4,748$   & $2,488$  & $1,426$  \\
\botrule
\end{tabular}
\end{minipage}
\end{center}
\end{table}

\subsubsection*{SAFE Project Multi-Species Multi-Task (SAFE-MSMT) Dataset}
\label{subsup:malaysia}

The raw dataset contains $120$ recording WAV files sampled at 16kHz, recorded in the context of the Stability of Altered Forest Ecosystems (SAFE) project \cite{ewers2011large}, and comprising $2$ hours, $24$ minutes, and $20$ seconds of audio, with $2,494.88$ seconds of annotated calls ($28.8\%$ of total duration).

We were able to identify and categorise calls belonging to $62$ different species. In three cases where an animal call was audible, but we were not able to identify the animal that produced it, we treated the corresponding audio duration as a negative call, comprising background sound.
 
The average call duration was $3.17 \pm 6.92$ seconds, however more than $80\%$ of the calls were shorter than $3$ seconds, with some outliers reaching durations of around $45$ seconds, so we opted for $3$-second clips to mimic the input shape in the setup of \cite{rizos2021multi} for the OSA-SMW dataset.

In an effort to define train-validation-test partitions that do not suffer from site-based, or temporal correlation bias, we performed partitioning in a recording-independent manner. We furthermore needed positive samples for each species category to be present per partition, and thus looked for species that are audible in at least three different recordings, that could be allocated into partitions in the desired manner. We identified $30$ classes that satisfied these desiderata.

Out of the total recording duration, $14.26\%$ of it contained overlapped calls. There were $29$ target species with at least one overlapped call (only the brown fulvetta, the third rarest species in the dataset, did not have any overlap). Due to this non-insignificant of overlapped calls, we decided to address this dataset through the multi-task framework.

\begin{figure}[t]%
\centering
\includegraphics[width=1.0\textwidth]{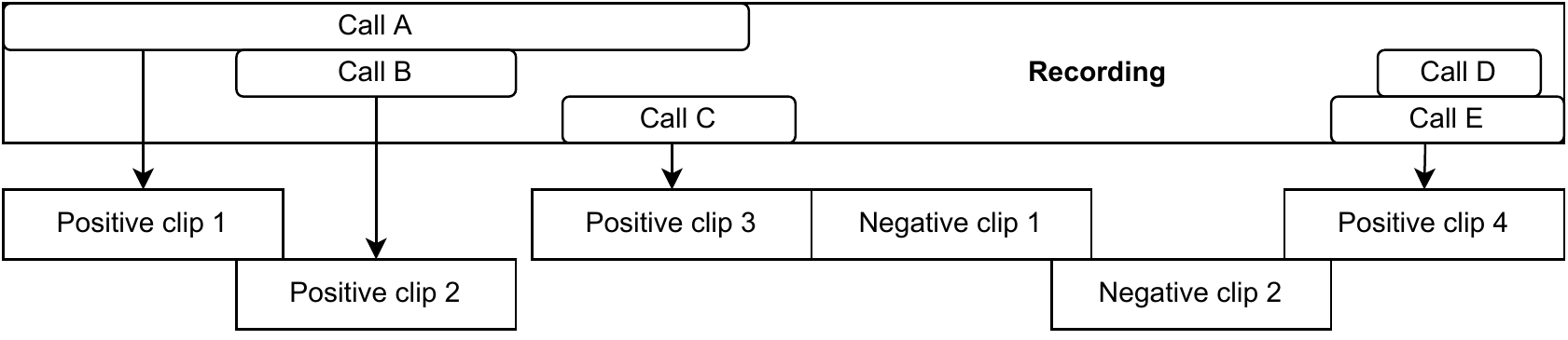}
\caption{Segmentation of original recording into $3$-second positive and negative clips. Clip 1 contains part of call A and call B; smaller than our threshold size for the latter, so we also extract clip 2, that contains the full call B, and part of call A. Clip 3 contains call the full call C, and part of call A. There is a short segment of call A (between clips 2 and 3) that is not included in any positive clip because it is shorter than our threshold. Calls D and E are both included in clip 4. Finally, the background noise part of the recording that does include any positive calls is chunked into two slightly overlapping negative clips.}
\label{fig:segment}
\end{figure}

In the segmentation of the original recordings into $3$ second clips, we try to ensure the inclusion of full calls in each clip, if they fit. It is possible that trying to do so for one call, one might fragment other partially overlapping calls which are not receiving focus. These calls are, however, revisited, such that for each call there is at least one clip that completely supports it, as depicted in Figure~\ref{fig:segment}. In the case where the calls are longer than $3$ seconds, we chunk them into (whenever possible) non-overlapping clips. For each such clip, we follow through with the clip extraction if there is a substantial portion of the call duration that was not supported by a previously extracted clip; \ie at least $1$ second.

Purely negative recordings are deterministically chunked into $3$ second clips that are negative for all classes/tasks. In recordings where identifiable calls exist, we similarly chunk the background sound segments between calls, if they are at least $3$ seconds long.

In order to instill randomness in the position of the call segment in the clip, the start time for each clip $t_{0}^{clip}$ was drawn from a uniform distribution $t_{0}^{clip} \sim \mathcal{U}[t_{T}^{call} - L ,t_{0}^{call}]$, where $t_{0}, t_{T}$ denote, respectively, the start and end time of the clip or the call, and $L$ the length of the clip; in our case three seconds. This ensures that short calls ($\leq L$) will be completely included in the clip, and also disassociates calls with any particular position in the clip. This step can avoid the model associate calls with a particular position in the clip.
 
We extracted $1,326$ negative and $613$ positive clips. $241$ of the latter contained one single call ($39.32\%$) while $368$ contained multiple complete or partial calls ($60.03\%$). The average number of calls in a positive clip was $2.06$, and the maximum number of calls in a clip was $7$.

In the end, each of these $30$ classes contains $4$ to $229$ call occurrence counts (chunked in case of long calls), for a total of $1,498$ calls in $519$ clips in the dataset.

\begin{figure}[t]%
\centering
\includegraphics[width=1.0\textwidth]{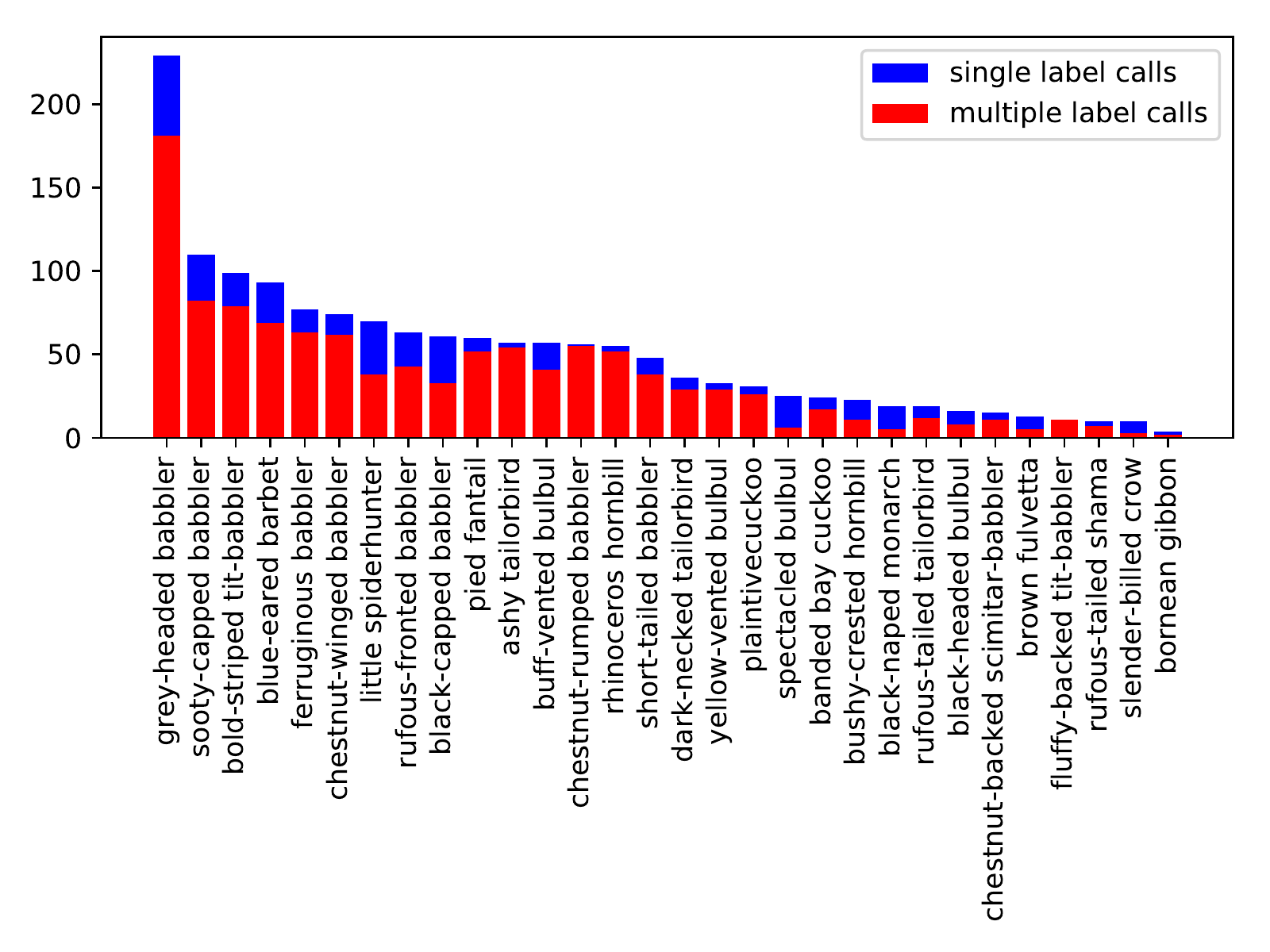}
\caption{This barchart summarises the total call count per species that we included in the study (after chunking long calls). The total height (in blue) denotes the total count, whereas the red part denotes the count per species that can be found in a clip with a call made by another species.}
\label{fig:overlap}
\end{figure}

Out of these calls, $1,124$ of them -- \ie $75.03\%$ of the total count -- coexist in clips with calls from other species. Figure~\ref{fig:overlap} depicts the overlapped duration against total call length for each target species. There are $435$ clips with multiple labels.

% \begin{figure}[!ht]
% \centering
% \includegraphics[width = 0.6\hsize]{./figures/callsperclip.png}
% \caption{Number of calls in each positive clip}
% \label{fig:callsperclip}
% \end{figure}

\subsubsection*{Osa Peninsula Spider Monkey Whinny (OSA-SMW) Dataset}
\label{subsup:costarica}

This dataset, first introduced in \cite{rizos2021multi}, resulted from a PAM study that covered approximately $1,000$ km$^2$ in the OSA peninsula of Costa Rica, managed by the Area de Concervaci{\'o}n Osa (ACOSA). Data from $13$ sites were included, recorded at a $48$ kHz sampling rate, and resampled to $16$ kHz for extraction of log-Mel spectrograms. The annotation focuses on the common 'whinny' call of A. Geoffroyi, which conveys general communication related to feeding and movement \cite{campbell2008spider}. The dataset contains $591$ positive whinny calls, each being around one second long, and we opted to segment our recordings into three second clips. We introduce randomness in the clipping of positive samples, as well as chunk the negative segments in a manner similar to how we described in \ref{subsup:malaysia} for the SAFE-MSMT dataset. We partition in a site-independent manner, to ensure that our model does not learn the acoustic characteristics of a site. Compared to \cite{rizos2021multi}, in this article we use one more test site.

\subsection*{Experimental Details}
\label{sup:experimentaldetails}

The architecture of our model, including numbers of filters and hidden units is depicted in Table~\ref{tab:seresnet}. We use a batch size of $8$, and the Adam optimiser, with starting learning rate of $10^{-5}$. We evaluated on the development set after the end of each training epoch, and stopped training with a patience -- \ie no improvement on (W-)AU-PR -- of $15$ epochs. We use $2$ SpecAugment time masks with a size of $24$ frames (\ie $240$ ms), and $2$ frequency masks with a size of $16$ log-Mel filters. After the masking, normally distributed additive jitter is used, with zero mean and a standard deviation of $10^{-6}$. The regularisation parameter (cold posterior) for the Bayesian KL loss factor is $10^{-10}$.

%\section{Per Species Performance in the Malaysia Multi-Species Multi-Task Dataset}
%\label{sup:perspeciesperformanceinthemalaysoiamultispeciesmultitaskdataset}




% \end{appendices}

% \bibliographystyle{unsrt}
% \bibliographystyle{unsrtnat}
\bibliography{sn-bibliography}

\end{document}